\pdfoutput=1

\documentclass[letterpaper, 10 pt, conference]{ieeeconf}
\IEEEoverridecommandlockouts 

\usepackage{amsmath}
\usepackage{amssymb}
\usepackage[nospace]{cite}
\usepackage{url}
\usepackage{xcolor}
\usepackage{fancyhdr}
\DeclareMathOperator*{\argmin}{\arg\!\min}
\usepackage{adjustbox}
\usepackage{graphicx}
\usepackage{tensor}
\usepackage[utf8]{inputenc}
\usepackage[OT1]{fontenc}
\usepackage[final]{microtype}
\usepackage{booktabs}
\usepackage[binary-units]{siunitx}
\usepackage{footmisc}

\makeatletter
\let\NAT@parse\undefined
\makeatother
\usepackage[hidelinks]{hyperref}

\title{\LARGE \bf
Learning-based Localizability Estimation for Robust LiDAR Localization
}

\author{Julian Nubert$^{\dagger1,2}$, Etienne Walther$^{\dagger1}$, Shehryar Khattak$^{1,2}$ and Marco Hutter$^{1}$
\thanks{This work is supported in part by the Max Planck ETH CLS, the EU Horizon 2020 programme grant agreement No.852044 and 101016970, the NCCR digital fabrication and robotics, and the SNSF project No.188596.}
\thanks{$\dagger$ Indicates equal contribution.}
\thanks{$^{1}$The authors are with the Robotic Systems Lab, ETH Z\"urich.} 
\thanks{$^{2}$The authors are with the Max Planck ETH CLS.}
\thanks{Corresponding Author: Julian Nubert, \tt\small nubertj@ethz.ch}%
}

\begin{document}

\maketitle
\thispagestyle{empty}
\pagestyle{empty}

\begin{abstract}
LiDAR-based localization and mapping is one of the core components in many modern robotic systems due to the direct integration of range and geometry, allowing for precise motion estimation and generation of high quality maps in real-time.
Yet, as a consequence of insufficient environmental constraints present in the scene, this dependence on geometry can result in localization failure, happening in self-symmetric surroundings such as tunnels.
This work addresses precisely this issue by proposing a neural network-based estimation approach for detecting (non-)localizability during robot operation. Special attention is given to the localizability of scan-to-scan registration, as it is a crucial component in many LiDAR odometry estimation pipelines.
In contrast to previous, mostly traditional detection approaches, the proposed method enables early detection of failure by estimating the localizability on raw sensor measurements without evaluating the underlying registration optimization.
Moreover, previous approaches remain limited in their ability to generalize across environments and sensor types, as heuristic-tuning of degeneracy detection thresholds is required.
The proposed approach avoids this problem by learning from a collection of different environments, allowing the network to function over various scenarios. Furthermore, the network is trained exclusively on simulated data, avoiding arduous data collection in challenging and degenerate, often hard-to-access, environments.
The presented method is tested during field experiments conducted across challenging environments and on two different sensor types without any modifications. The observed detection performance is on par with state-of-the-art methods \emph{after} environment-specific threshold tuning\footnote{Supplementary Video: \url{https://youtu.be/fm08PFwMO0c}}.
\end{abstract}

\section{Introduction}\label{sec:intro}
Light Detection And Ranging (LiDAR) sensors are one of the most common sensors used for robot localization and mapping to date. The ability to provide depth measurements at long ranges and to operate under varying illumination conditions along with recent miniaturizing and reduction in costs have significantly increased their suitability for various robotic applications. To estimate robot pose from LiDAR data, popular techniques iteratively minimize the distance between point- or feature-correspondences in consecutive LiDAR scans to obtain the relative scan-to-scan pose transformations~\cite{gicp,ndt}. Furthermore, the individual LiDAR scans are aligned and merged to create a map representation of the environment, which facilitates further robot pose refinement and accurate map creation through scan-to-map~\cite{loam} or scan-to-representation registration~\cite{suma}. However, in order to converge to the correct solution these algorithms depend on sufficient geometric constraints in the environment. In the absence of such constraints, as in  geometrically self-similar or symmetric environments, the optimization problem can become ill-conditioned and provide a sub-optimal solution. While the trend towards learned scan-to-scan estimation approaches~\cite{nubert2021self,lonet} can redress the need of hand-crafting features, it can not eliminate the occurrence of such environment dependent "LiDAR slip", often occuring in corridors, hallways, tunnels or planar environments.

\begin{figure}[t!]
\setlength{\lineskip}{0pt}
\centering
\includegraphics[width=\columnwidth]{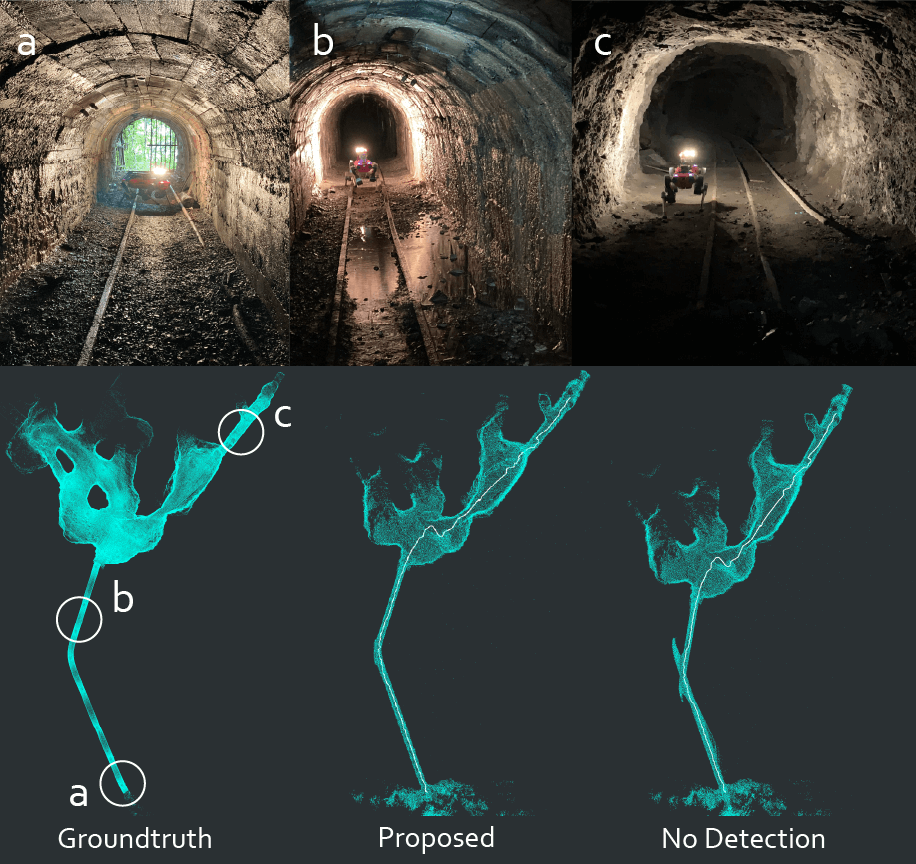}
\caption{Snapshots and maps from a challenging field test in an underground mine in Switzerland. The top-row images showcase different segments along the path that are $a)$ localizable, $b)$ non-localizable and $c)$ mostly localizable. The bottom row shows the created map with and without the proposed localizability detection.}
\label{fig:main}
\vspace{-2ex}
\end{figure}

To mitigate such problems, most of the techniques focus on the exposure of ill-conditioned optimization problems. However, they remain limited in two aspects: first, most techniques make use of eigenvalues or the condition number of the approximate hessian matrix of the optimization problem. This,  however,  requires computationally expensive steps of data filtering and establishing point correspondences to be performed before the detection. Second, the detection of optimization degeneracy mostly relies on a heuristic threshold, which can vary significantly between different environments and sensors, hence limiting robot operation in degenerate environments without manual re-tuning of thresholds. 
Furthermore, it should be noted that most existing methods that detect optimization degeneracy are dependent on both the target and reference point cloud scans and cannot evaluate the data quality provided by a single scan.

To overcome these challenges, this work proposes a method for learning-based localizability estimation from a single point cloud scan. The predicted metric encapsulates how localize-able a given point cloud would be w.r.t other point cloud scans, and can be used to predict upfront whether scan-to-scan registration will succeed for $3D$ pose estimation. Furthermore, the proposed approach is trained in an end-to-end manner using simulated sensor data only, and hence, avoids data collection in degenerate and often difficult-to-access environments. The experiments performed both in simulation and on real-world data demonstrate that the learned-approach can correctly predict localizability across a variety of environments without a need of re-tuning, is capable to work with real data, and can even generalize to different LiDAR types.
The main contributions of this work are as follows: $i)$ The development of a learning-based approach that successfully predicts the LiDAR localizability in $6$-DOF directly from point cloud data. $ii)$ A thorough evaluation conducted to improve the robot localization performance of the ANYmal-C quadruped robot operating in challenging and degenerate real-world environments. $iii)$ A thorough design and implementation of all components, including the used dataset and dataset generation. All relevant parts will be made publicly available as an open-source package for the benefit of the robotics community\footnote{\label{footnote:github}\url{https://github.com/leggedrobotics/L3E}}.

\section{Related Work}\label{sec:related}
Pose uncertainty estimation is an essential component of the robot state-estimation process, as it not only provides a quantitative measure of the quality of the pose estimates provided by a sensing modality, but also facilitates reliable sensor fusion. For LiDAR-based pose estimation, the works in~\cite{censi2007_1,censi2007_2} propose a closed-form estimate for Iterative Closest Point (ICP) methods by carefully analyzing the effect of incorrect convergence, under-constrained situations, and sensor noise on the error function. However, as discussed in~\cite{bonnabel2016covariance}, this closed-form formulation tends to be over-optimistic with regard to that pose uncertainty, and building on this insight, the authors in~\cite{brossard2020} propose to not ignore the effect of initialization accuracy on the final pose uncertainty. 
To avoid overconfidence and to include all potential sources of errors, multiple methods~\cite{icp_uncertainity_monte_carlo_sim_2017,bayesionICP2020} employ sampling approaches to achieve a better ICP pose uncertainty estimate at the expense of additional computation. These formulations try to include all potential sources of errors under a single quality metric, which can be beneficial for sensor fusion tasks. However, such approaches do not provide an independent understanding of the quality of geometric constraints provided by the environment, which may be the primary cause of optimization ill-conditioning in challenging environments.

\begin{figure*}[t!]
\includegraphics[width=\textwidth]{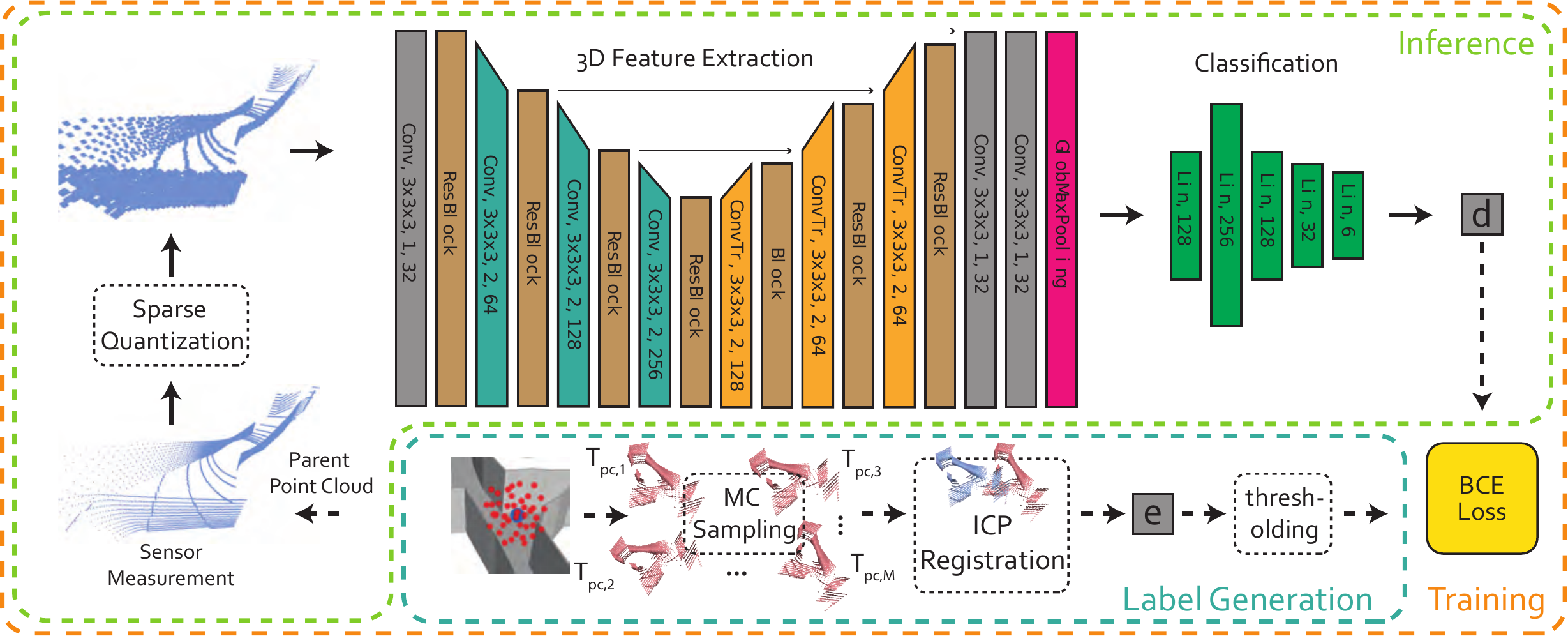}
\caption{Overview of the proposed approach. The input point cloud received from the LiDAR sensor is down-sampled, sparsely quantized, and then fed into a sparse $3D$ convolutional neural network based on the ResUnet architecture. Monte Carlo sampling and ICP point cloud registration are used to generate the ground truth labels required for the supervision signal in order to train the feature extraction and classification network in an end-to-end fashion.}
\label{fig:overview}
\vspace{-2ex}
\end{figure*}

To detect optimization degeneracy in cases where environment do not provide enough geometric constraints,~\cite{zhang2016degeneracy} proposes a method relying on eigenvalues to detect ill-conditioning and proposes to perform solution remapping along the degenerate directions. This approach has proven to perform well in various real world scenarios, e.g. in~\cite{tranzatto2022cerberus}, however, the performance of the approach relies on heuristically defined eigenvalue thresholds that are highly dependent on the environment and the deployed sensor suite. 
Similarly,~\cite{kaess2019degeneracy} relies on the condition number to determine the health of the optimization process and includes partial constraints along non-degenerate direction for sensor fusion. Other methods, such as~\cite{NDTclassify2018,correctAligned2021} rely on the final alignment of scans to capture the adequacy of geometric constraints provided by the environment for correct solution convergence. However, all these methods either rely on the point cloud registration process or its result to determine the performance of the pose estimation process, and do \emph{not} exploit the information provided by the point cloud data directly to facilitate the estimation process itself. 
Addressing this issue,~\cite{zhen_robust_2017,zhen_estimating_2019} propose to quantify the geometric constraints provided by point cloud scan or map data and suggest to focus on a localizability metric instead of degeneracy detection. In~\cite{zhen_robust_2017}, the authors demonstrate measuring localizability of a point cloud map by evaluating constraints provided by the data, while in~\cite{zhen_estimating_2019} they use a similar approach to perform sensor fusion when insufficient constraints are present during the traversal along a tunnel-like environment. In a similar manner,~\cite{liu_localizability_2021} proposes to segment the map first into geometric features and then evaluates whether the combined constraints provided by all features are sufficient for determining the 6-DOF pose accurately. However, one of the main limitations of all these methods is the determination of a heuristic threshold that is highly dependent on the environment and difficult to pre-determine in a reliable manner. 

Arguing along similar lines,~\cite{cello} proposes to employ machine learning techniques to predict robot pose uncertainty by directly learning from sensor data. The proposed technique demonstrates learning-based $2D$ pose covariance prediction by utilizing planar LiDAR features. Extending the same concept to $3D$, authors in ~\cite{cello3d} demonstrate sensor data-driven learning methods for scan-to-scan matching of $3D$ LiDARs. Although these methods demonstrate the heuristic-free advantage of learning-based methods, they still rely on careful extraction and selection of features from sensor data, rendering them sensitive to the feature selection process and limit the utilization of all available sensor data. 
To utilize the raw sensor outputs,~\cite{pointnetkl2020} demonstrates the application of an end-to-end deep-learning method to predict the pose estimation uncertainty of an autonomous underwater vehicle directly from $3D$ bathymetric point clouds. However, the given method is limited to $2D$ robot navigation in specific underwater scenarios, leaving a gap to higher dimensional and more complex robotic systems.

Given the discussion above, this work fills a gap by proposing a heuristic-free learning-based 6-DOF localizability estimation approach. The proposed approach is capable of operating directly on LiDAR scan data in an end-to-end manner and avoids the need for developing hand-crafted features. The proposed technique is fully trained in simulation, and avoids the difficult task of data collection in challenging geometrically degenerate environments, and is capable to generalize to a variety of real-world environments without tuning or modification.

\section{Problem Formulation}\label{sec:problem}
\label{sec:problem_formulation}
Given the aforementioned discussion, the goal of this work is to reliably predict the ability to successfully localize in certain directions using scan-to-scan registration in the current environment. For this purpose, this work aims to provide a localizability estimate directly from a $3D$ LiDAR point cloud scan $\textbf{s}_{k} \in \mathbb{R}^{n_k \times 3}$. Conceptually, localizability can be thought of as a detection measure along 6-DOF, which in this work is defined as 
\begin{equation}
\textbf{d}_k = \left( d_x, d_y, d_z, d_\phi, d_\theta, d_\psi \right)^\top,
\end{equation}
with $x$, $y$, $z$ denoting translation coordinates, and $\phi$, $\theta$ and $\psi$ denoting the Euler angle representation for roll, pitch and yaw, respectively. For practical reasons, the localizability is considered binary; either a certain direction is localizable or not. Hence, it holds $d_i \in \{0, 1\}~\forall~ i\in \{x,y,z,\phi,\theta,\psi\}$, where $0$ (localizable) and $1$ (non-localizable) express whether or not the given direction can be localized successfully in the current environmental surroundings via scan-to-scan LiDAR registration of $\mathbf{s}_k$ against scan $\mathbf{s}_{k-1}$.

The estimation of $\mathbf{d}_k$ is formulated as a multi-label binary classification problem, and is performed through a neural network classifier. The localizability at time $k$ is modelled to be fully determined by the current point cloud scan observation $\mathbf{s}_k \in \mathbb{R}^{n_k \times 3}$. The resulting function $\mathbf{s}_k \rightarrow \mathbf{d}_k$ can hence be approximated through $\Tilde{\mathbf{d}}_k(\Theta,\mathbf{s}_k)$, where $\Theta \in \mathbb{R}^P$ are the $P$ trainable parameters of the network. These parameters are obtained via minimization of a supervised classification loss $\argmin\limits_{\Theta} \mathcal{L}(\Tilde{\mathbf{d}}(\Theta,\mathcal{S}), \mathcal{T})$ over a training set of scans $\mathbf{s}_i \in \mathcal{S}$ with ground truth labels $\mathbf{t}_i \in \mathcal{T}$.

\section{Proposed Approach}\label{sec:approach}
As introduced in Section~\ref{sec:problem_formulation}, the goal of this work is to reliably determine the localizability measure $\mathbf{d}_k$ at time $k$ during robot operation by only considering the current LiDAR scan $\mathbf{s}_k$. This section details the proposed approach. An overview of the steps taken is shown in  Figure~\ref{fig:overview}.

\subsection{Localizability Measure}
\label{sec:degen_meas}
The specific definition of the localizability measure $\mathbf{d}_k$, e.g. needed for the training-data generation, requires two steps: first an expected registration error $\mathbf{e}_k \in \mathbb{R}^6$ is computed, similarly to~\cite{pointnetkl2020}.  Second, this registration error is mapped to the desired metric $\mathbf{d}_k$ through a thresholding operation. 

\subsubsection{Expected Registration Error}
In order to compute the expected registration error $\mathbf{e}_\mathtt{p}$ for a \emph{parent point cloud} $\mathbf{s}_\mathtt{p}$, the average registration residual of a defined point cloud distribution around $\mathbf{s}_\mathtt{p}$ needs to be determined.

Given $\mathbf{s}_\mathtt{p}$, the first step is to perform Monte Carlo sampling for a collection of $M$ \emph{child point cloud} scans $\mathbf{s}_{\mathtt{c},j}, j\in\{1,...,M\}$ in proximity of the parent cloud.
To do so, the required child point cloud poses $\mathbf{T}_{\mathtt{p},\mathtt{c},j} \in SE(3)$ are computed by drawing 6 perturbations in directions $i \in \{x,y,z,\phi,\theta,\psi\}$ from zero-mean Gaussians $\mathcal{N}(0,\sigma_i^2)$ with parameters given in Table~\ref{table:sampling_params}, and converting them to $SE(3)$. The selected sampling parameters are chosen according to the maximum expected motion of the robot between two consecutive scans.
Each resulting child point cloud is then registered against the parent point cloud using point-to-plane ICP. The expected registration error $\mathbf{e}_\mathtt{p}$ for parent point cloud $\mathbf{s}_\mathtt{p}$ is computed as the mean absolute error:
\begin{equation}
\label{equ:registration_error}
    \mathbf{e}_\mathtt{p} = \frac{1}{M} \sum_{j=1}^M \bigg| \gamma \left( \Tilde{\mathbf{T}}_{\mathtt{p}\mathtt{c},j}^{-1} \cdot \mathbf{T}_{\mathtt{p}\mathtt{c},j} \right) \bigg|.
\end{equation}
Here, $\Tilde{\mathbf{T}}_{\mathtt{p}\mathtt{c},j} \in SE(3)$ denotes the estimated transformation from the registration algorithm, and $\gamma$ denotes the projection back from $SE(3)$ into $(x,y,z,\phi,\theta,\psi)$.
\begin{table}[h!]
    \centering
    \caption{Parameters chosen for performing the Monte Carlo sampling.}
    \begin{tabular}{c c c c c c c} 
         \toprule
          $M$ & $\sigma_x$[m] & $\sigma_y$[m] & $\sigma_z$[m] & $\sigma_\phi$[°] & $\sigma_\theta$[°] & $\sigma_\psi$[°] \\ 
          \midrule
          200 & 0.1 & 0.1 & 0.1 & 5.0 & 5.0 & 10.0 \\
          \bottomrule
    \end{tabular}
    \label{table:sampling_params}
\end{table}
Note, that due to the sparse and circular structure of LiDAR scans, in contrast to the approach in~\cite{pointnetkl2020}, scans $\mathbf{s}_\mathtt{p}$ and $\mathbf{s}_{\mathtt{c},j}$ can not be the same perturbed scan. Instead, an additional point cloud scan must be sampled at every child pose. 

\subsubsection{Obtaining Localizability Measure}
After computing the $6D$ registration errors, the binary labels in $\mathbf{d}_k$ can be obtained by applying thresholds to the expected registration error:
\begin{equation}
    \mathbf{d}_k = \alpha(\mathbf{e}_k).
\end{equation}
Here, $\alpha$ is a component-wise thresholding operation, with the motion thresholds as presented in Table~\ref{table:thresholds}.
\begin{table}[ht!]
    \centering
    \caption{Motion thresholds for computing $\mathbf{d}_k$ from $\mathbf{e}_k$.}
    \begin{tabular}{c c c c c c} 
         \toprule
          $e_x$[m] & $e_y$[m] & $e_z$[m] & $e_\phi$[°] & $e_\theta$[°] & $e_\psi$[°] \\ 
          \midrule
          0.1 & 0.1 & 0.1 & 2.0 & 2.0 & 2.0 \\
          \bottomrule
    \end{tabular}
    \label{table:thresholds}
\end{table}
These values are selected according to the sampling parameters from Table~\ref{table:sampling_params}, considering a less than $10$cm error in translation and $2$° in rotation as indication of convergence.

\subsection{Architecture}
The choice of the right architecture and data representation is one of the key building blocks when developing a neural network estimation method. 

\subsubsection{Data Representation \& Feature Extraction}
Compared to neural network architectures deployed for image processing, with $2D$ convolutional neural networks being the prominent choice, the selection of the best network architecture for processing $3D$ point cloud data is often an open question.
Commonly chosen architectures include but are not limited to: set-based methods~\cite{pointnet}, graph-based approaches~\cite{shi2020point}, image-view techniques, and voxelization-based architectures~\cite{voxnet}. While projection-based image-view representations have been chosen frequently for rotating LiDAR sensors~\cite{nubert2021self,lonet,stanislas2021airborne} due to low memory requirements and the possibility to use well-explored $2D$ convolutional neural networks, generalization to different sensor types with varying field of view, intensity characteristics, or number of vertical rays remains difficult.
In a related work~\cite{pointnetkl2020}, the authors propose to make use of the PointNet architecture to predict $2D$ localization covariances. While these set-based methods have shown to perform well for global classification tasks, they were specifically designed to be invariant to rigid body transformations.
Often deemed to be memory hungry and inefficient, in recent years voxel-based $3D$ convolutional neural networks gained importance through the rise of sparse convolutions~\cite{minkowski2019}.
Motivated by these developments, this work builds on an architecture based on sparse $3D$ convolutions, allowing the network to develop a good scene understanding while remaining fast and memory efficient. To justify this network choice, a comparison against the PointNet architecture for the localizability prediction task is presented in Section~\ref{sec:validation}.

A full overview of the utilized network architecture is presented in Figure~\ref{fig:overview}. The network design is based on the $3D$ ResUNet~\cite{resUNet2018} architecture and is inspired and adopted from~\cite{fcgf2019}, within which state-of-the-art results for point cloud feature learning are demonstrated.
Given a LiDAR point cloud provided as an input, first a sparse quantization is performed to down-sample the point cloud to the correct density and to obtain the correct data-format. The architecture is generally built up like a UNet but with a ResNet block consisting of two convolutional networks plus a skip connection after each down- and up-sampling convolutional layer. Global max pooling is applied at the end in order to get a constant-size feature vector of size 32.

\subsubsection{Classification Network}
The output of the previous feature extracting network is then forwarded to an MLP with 5 layers. The final output of the network is a $6D$ vector. After channeling each of the entries through a sigmoid activation function, a probability estimate $\Tilde{\mathbf{p}_k}$ is obtained.

\subsection{Loss}
Formulating the objective as a classification problem circumvents addressing, in part, the very different scales of translation and rotation components.
Binary cross entropy is chosen to be the loss function during training, which for the raw probability vector $\Tilde{\mathbf{p}}_k$ is defined as
\begin{equation}
    \mathcal{L}_k = - \frac{1}{6} \sum_{i=1}^6 t_{k,i} \cdot \text{log}(p_{k,i}) + (1 - t_{k,i}) \cdot \text{log}(1 - p_{k,i}).
\end{equation}
Here, $t_{k,i}$ denotes the target label obtained during the procedure described before.

\subsection{Classification Threshold}
Instead of dealing with hard-to-interpret eigenvalue metrics, which often require manual tuning of detection thresholds, the output of the proposed network has the characteristics of a probability estimate for each of the classes.
\begin{table}[ht!]
    \centering
    \caption{Probability thresholds for determining $d_{k,i}$ from $p_{k,i}$.}
    \begin{tabular}{c c c c c c} 
         \toprule
          $p_x$[m] & $p_y$[m] & $p_z$[m] & $p_\phi$[°] & $p_\theta$[°] & $p_\psi$[°] \\ 
          \midrule
          0.3 & 0.3 & 0.8 & 0.3 & 0.8 & 0.8\\
          \bottomrule
    \end{tabular}
    \label{table:prob_thresholds}
\end{table}
To transform the provided probabilities into class labels, a natural but na\"ive choice is to threshold each of the entries of $\Tilde{\mathbf{p}}_{k,i}$ at $50$\%. Instead, in this work the thresholds are chosen according to a \emph{precision-recall} curve generated for the validation set. The same thresholds from Table~\ref{table:prob_thresholds} are used for all experiments within this work.

\subsection{Training Procedure}
For efficient training of the proposed network, some additional design consideration are taken into account.
\subsubsection{Dataset Balancing}\label{sec:balancing}
Although care has been taken during the dataset generation, the dataset is still highly imbalanced with regard to positive and negative instances for each of the six directions. In particular, there are very few instances in the dataset that are non-localizable for $z$, $\phi$ and $\theta$. 
To reduce this imbalance, so called \emph{power-set labels}~\cite{powerset} are used to balance the dataset. Overall, $2^6 = 64$ such labels are generated, where up-sampling is performed with respect to the second most common class. The maximum up-sampling factor is set to be $100$ in order to limit the over-sampling of heavily underrepresented power-set-labels. 

\subsubsection{Training Characteristics}
For training the network, stochastic gradient descent (SGD) with a batch-size of 8 is selected. The learning rate is chosen to decay at an exponential rate of $\gamma = 0.9$. To avoid over-fitting, dropout is utilized in the classification network. Further, 4000 random points in the point cloud scan are sampled during training and deployment, followed by sparse quantization with a voxel-size of $0.2$m.

\section{Data Generation \& Implementation}\label{sec:implementation}
This section details the environment generation and practical data sampling procedure according to the mathematical foundations laid out in Section~\ref{sec:degen_meas}. Furthermore, implementation details of the network are provided.

\subsection{Dataset Generation}
As introduced in Section~\ref{sec:approach}, the ground truth labels are generated via Monte Carlo sampling of child point clouds in a neighborhood of parent point clouds. 
In order to obtain meaningful scan-to-scan registration results that can be used for computing the expected registration error $\mathbf{e}_k$, more than just rotating and translating the parent point cloud is required to obtain suitable child point clouds. 
Instead, both source and target, i.e. parent and child point cloud, have to be recorded through ray-casting at the sampled sensor location (in simulation or reality). 
To render this possible, and to coincide with the theoretical sampling procedure introduced earlier in Section~\ref{sec:degen_meas}, this work performs the whole scan collection purely in simulation with the LiDAR sensor being spawned at parent and child poses.

\subsubsection{Environments}
\begin{figure}[t!]
\includegraphics[width=\columnwidth]{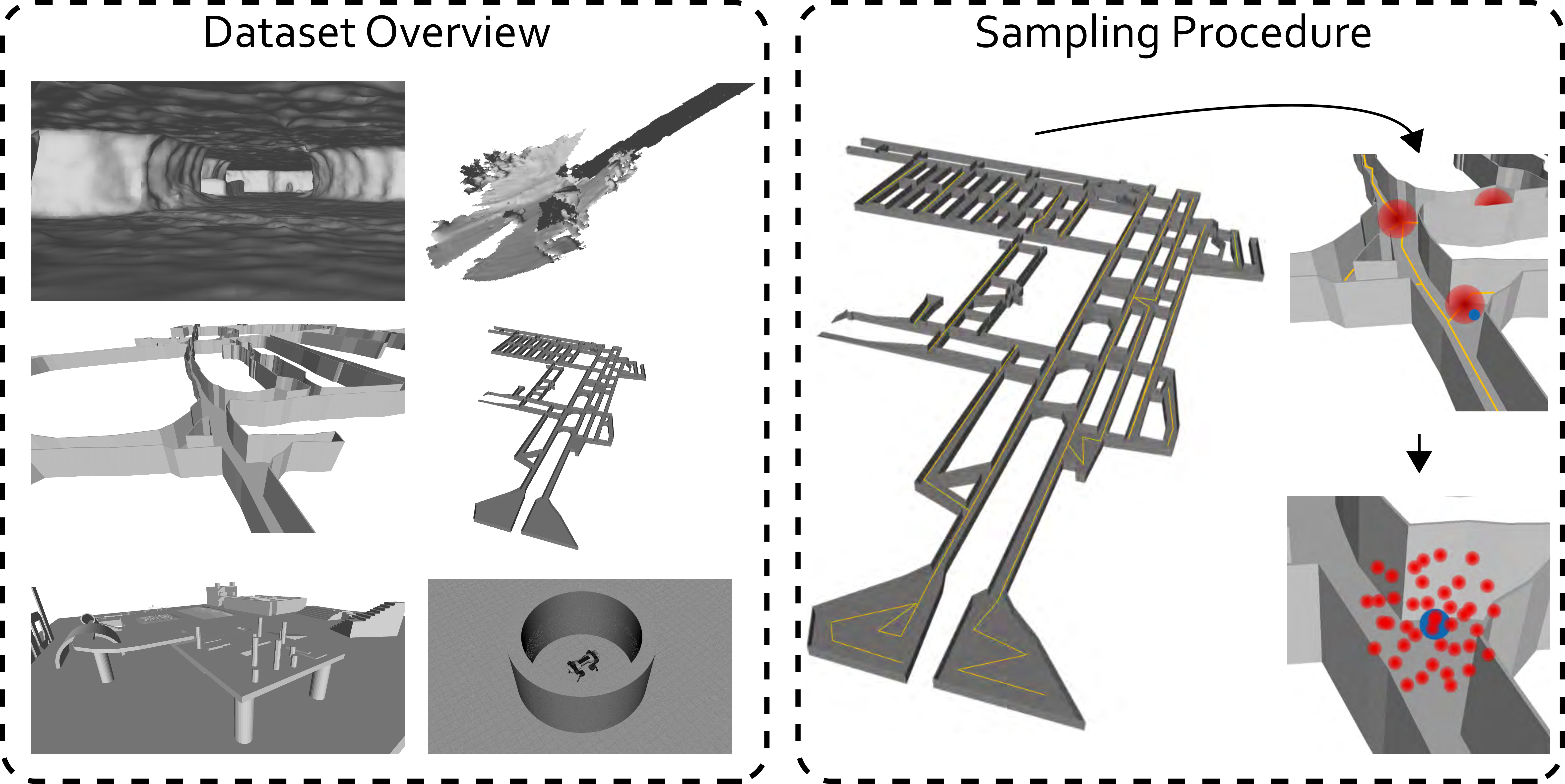}
\caption{Left: an overview of some of the used environments is shown, mainly created in CAD or sampled from underground environments. Right: an exemplary sampling path is depicted for collecting parent point clouds.}
\label{fig:dataset}
\vspace{-2ex}
\end{figure}

All point clouds are sampled in simulation from a collection of environments. The environment selection and generation is done in a way to represent as many of the basic types of degenerate and non-degenerate cases as possible. As a result, the created dataset contains instances of planes, tunnels, cylinders, and meshed cave environments. A small collection of these environments is shown in Figure~\ref{fig:dataset} on the left. Overall, 15 environments were collected for the final training and validation dataset generation. The used meshes are either obtained purely from CAD or by meshing down-sampled point cloud scans of real cave environments. All environments are made publicly available\footref{footnote:github}.

\subsubsection{Point Cloud Sampling in Simulation}
To sample a diverse set of parent point clouds, three measures are taken during the sampling process. 
First, in order to efficiently sample meaningful point clouds from inside the environments, \emph{sampling paths} consisting of poly-lines are hand-drawn along the ground of the meshes using Cloud Compare\footnote{\url{https://www.cloudcompare.org/main.html}}. An example of such a sampling path is shown in yellow in Figure~\ref{fig:dataset} on the right. Second, the parent point cloud sampling is performed in proximity to the sampling paths in Gazebo, with the point clouds being recorded using the simulated model of a Velodyne VLP-16 LiDAR. Each sensor pose in the world frame $T_\mathtt{wp}$ is sampled around selected points of the sampling path according to normal distributions with zero mean and the following variances: $\sigma_{\text{trans},x,y} = 0.2$m, $\sigma_{\text{trans},z} = 0.4$m, $\sigma_{\text{rot},\phi,\theta} = 15$°, $\sigma_{\text{rot},\psi} = 180$°. Third, collision checking for every resulting transformation $T_\mathtt{wp}$ is performed, and if collision-free, the child point locations are sampled around the parent pose as described in Section~\ref{sec:degen_meas}.

\subsection{Network Implementation}
The $3D$ convolutional neural network is implemented using PyTorch and MinkowskiEngine~\cite{minkowski2019}. The deployed network can be trained on a Nvidia RTX3090 GPU in about 4 hours. The model which is deployed during the experiments in Section~\ref{sec:experiments} was trained for 50 epochs. Without any specific code optimization, the inference time of the PyTorch model embedded into a ROS node is $28$ milliseconds for CPU-only mode (Intel i7 10700F) and $13$ milliseconds when being deployed on a Nvidia RTX3070 GPU. 

\section{Experimental Results}\label{sec:experiments}
To evaluate the feasibility of the proposed approach and to demonstrate its suitability towards real-world applications, a set of experimental studies are conducted.
First, the choice of the network architecture is validated through an ablation study. 
Next, the potential of the proposed approach is demonstrated through robotic field experiments conducted in three different environments, with the first two being highly degenerate.
Last, the generalization of the presented method is demonstrated by showing that the same trained network can work across two different LiDAR sensors, varying in their field-of-view and point cloud density, leading to similar performance when being deployed in the same environment.

\subsection{Network Architecture Validation}
\label{sec:validation}
\begin{figure}[t!]
\centering
\includegraphics[width=\columnwidth]{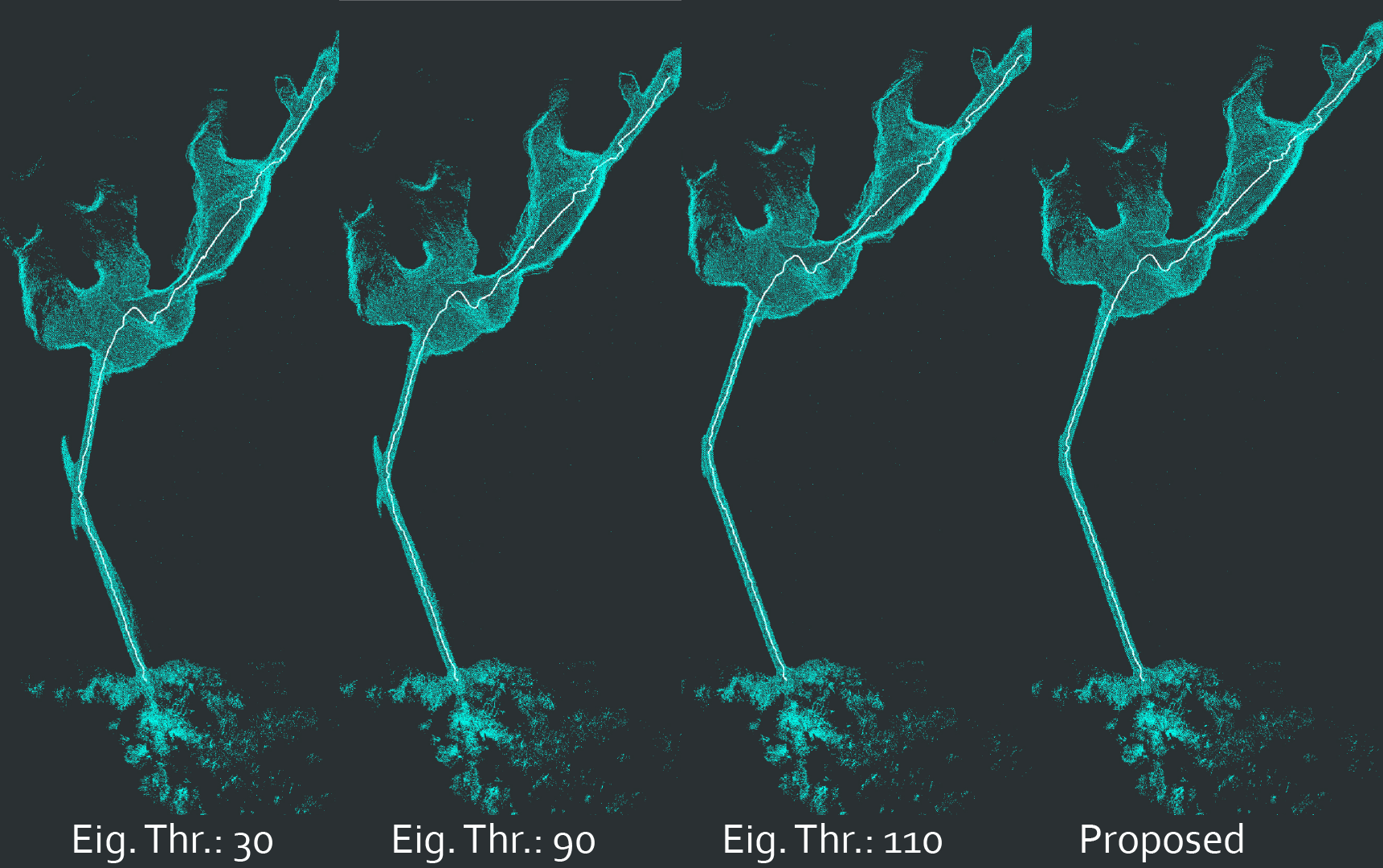}
\adjustbox{trim=0 {0.08\height} 0 0,clip, width=\columnwidth}
{\includegraphics[width=\columnwidth]{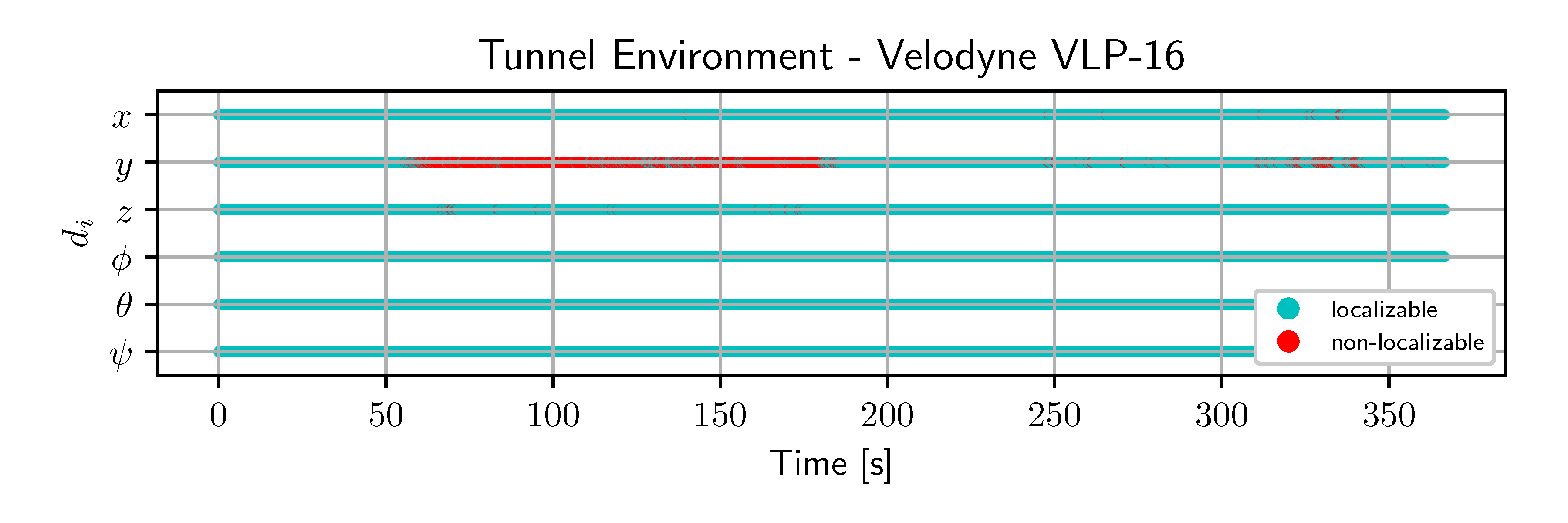}}
\caption{Predicted localizability and mapping results for the tunnel environment. Maps for three different eigenvalue thresholds $\{30,90,110\}$ as well as the proposed approach are presented. While an e-value threshold of at least 110 is required to reliably detect degeneracy, the network detects the degeneracy along the robot's $y$-axis without any intervention.}
\label{fig:seemuehle_maps}
\vspace{-2ex}
\end{figure}

To verify the suitability of using ResUNet over PointNet (as is used in~\cite{pointnetkl2020}), the architectures' performance is studied using the same training, validation and test sets. 
To that end, the performance on the training set is assessed for a better understanding of the expressiveness of the network, while the validation set is studied to assess the actual learning capability in familiar but unseen scenarios. 
Furthermore, a test set is created in simulation from a ground truth mesh of the tunnel environment (cf. Figure~\ref{fig:main},~\ref{fig:seemuehle_maps}), which is used to numerically compare the generalization abilities of the two models. The corresponding ground truth map was recorded using \emph{Leica RTC360} and \emph{BLK2GO} sensors.
Table~\ref{table:validation} provides insights into the performance of the two architectures trained for $60$ epochs, across the three described scenarios by listing accuracy, F1-score, precision, and recall. The shown values are the averaged values of the six dimensions $\{x,y,z,\phi,\theta,\psi\}$.

\begin{table}[h!]
    \centering
    \caption{Comparison of ResUNet and PointNet feature extractors. The shown values are averaged values over all six dimensionss.}
    \begin{tabular}{c|ccc|ccc} 
         \toprule
           & \multicolumn{3}{c}{ResUNet (proposed)} & \multicolumn{3}{c}{PointNet} \\ 
          \midrule
          & Train & Valid & Test & Train & Valid & Test \\ 
          Accuracy & $\mathbf{0.998}$ & $\mathbf{0.961}$ & $\mathbf{0.854}$ & $0.957$ & $0.945$ & $0.809$ \\ 
          F1-score & $\mathbf{0.997}$ & $\mathbf{0.606}$ & $\mathbf{0.517}$ & $0.94$ & $0.53$ & $0.214$ \\ 
          Precision & $\mathbf{0.997}$ & $\mathbf{0.647}$ & $\mathbf{0.398}$ & $0.918$ & $0.472$ & $0.22$ \\
          Recall & $\mathbf{0.998}$ & $0.585$ & $\mathbf{0.752}$ & $0.959$ & $\mathbf{0.703}$ & $0.244$ \\
          \bottomrule
    \end{tabular}
    \label{table:validation}
    \vspace{-1ex}
\end{table}

It can be noted that the ResUnet architecture outperforms PointNet in all evaluation scenarios except the recall on the validation set. The much better performance on the training set indicates more capacity to represent certain distributions. While this can potentially result in over-fitting, ResUNet also outperforms PointNet on the validation set. Finally, the performance of the proposed architecture is superior on the test set; when considering F1 score, precision and recall, the practical advantage of ResUNet over PointNet is underlined. Note, that the numeric result of precision and recall are higher for $x,y,\psi$, while being relatively low for $z,\phi,\theta$ as a consequence of few positive examples in the training set.

\subsection{Field Experiments}\label{sec:field_exp}
To evaluate the real-world performance of the proposed method, a set of field deployments were conducted using the ANYmal quadrupedal robot~\cite{fankhauser2018anymal} equipped with a Velodyne VLP-16 LiDAR. 
The field tests were conducted in: $i)$ a tunnel environment located in an underground mine, $ii)$ an open field at a paved work site, and $iii)$ an indoor urban environment containing narrow corridors and open spaces, offering a variety of challenging scenarios for LiDAR-based localization due to self-symmetry and lack of geometric constraints along different directions. 

To demonstrate the contribution of the proposed approach towards improving robot localization for operation in challenging environments, the localizability predictions  of the proposed approach are integrated as a degeneracy source into a complementary multi-modal localization and mapping framework, CompSLAM~\cite{compSLAM2020}. This framework was used by team CERBERUS~\cite{tranzatto2022cerberus} during their winning run of the DARPA Subterranean challenge, and is capable of handling single estimation failures through multi-modal sensor fusion. For the purpose of this work, the proposed method replaces the LiDAR degeneracy detection module in CompSLAM, which is originally based on the work of~\cite{zhang2016degeneracy}. Hence, in-addition to evaluating the impact on real-world robot localization performance, this also allows for a comparison against current state-of-the-art LiDAR degeneracy detection methods.

\subsubsection{Tunnel Environment}\label{sec:tunnel_env}
To evaluate the performance of the proposed method in a real-world LiDAR degenerate environment, a field test was conducted at Seem\"uhle mine, Switzerland. This abandoned mine has a long tunnel corridor connecting the entrance to the main quarry area as shown in the ground truth scan in Figure~\ref{fig:main}. During this experiment, the ANYmal-C robot started outside the mine and traversed towards the quarry. Both these areas have enough geometrical features for LiDAR-based localization to function properly, however, the connecting tunnel has smooth walls and is symmetric along its principal axis, representing a practical failure scenario. Accurate prediction of localizability is essential for robot operation in this environment to provide reliable robot localization through sensor-fusion. In this experiment, kinematic leg-odometry based on work of~\cite{tsif} is fused with LiDAR localization in CompSLAM along the predicted non-localizable DOF, in order to avoid localization failure and to produce an accurate map of the environment, as shown in Figure~\ref{fig:main},~\ref{fig:seemuehle_maps}. It should be noted that the proposed method is not trained in this environment, but only on sampled generated in simulation.
Furthermore, to demonstrate the advantage of the heuristic-tuning-free nature of the proposed method, the deterioration effects of using incorrect heuristic thresholds on robot localization and mapping in such challenging environments is shown in the top of Figure~\ref{fig:seemuehle_maps}. Different eigenvalue thresholds are set for the default degeneracy detection approach~\cite{zhang2016degeneracy} used in CompSLAM and it can be observed that the map quality is sensitive to even small changes in the thresholds used, corroborating the need for heuristic-free approaches. In contrast, the lower of Figure~\ref{fig:seemuehle_maps} shows the proposed approach detecting the degeneracy along the robot's y-axis correctly, which allows for a non-corrupted map creation without parameter tuning.

\subsubsection{Open Field}
\begin{figure}[t!]
\includegraphics[width=\columnwidth]{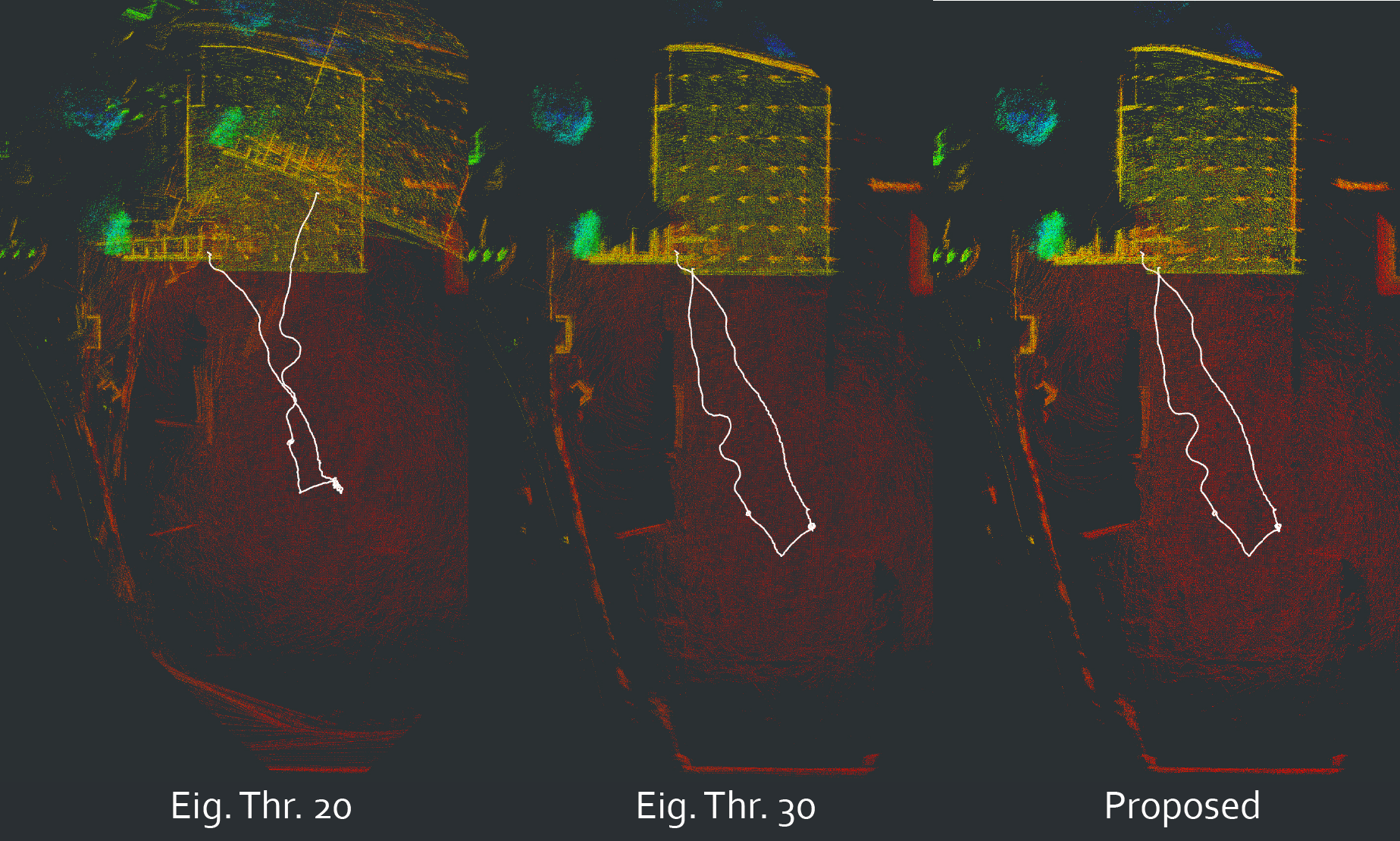}
\adjustbox{trim=0 {0.08\height} 0 0,clip, width=\columnwidth}
{\includegraphics[width=\columnwidth]{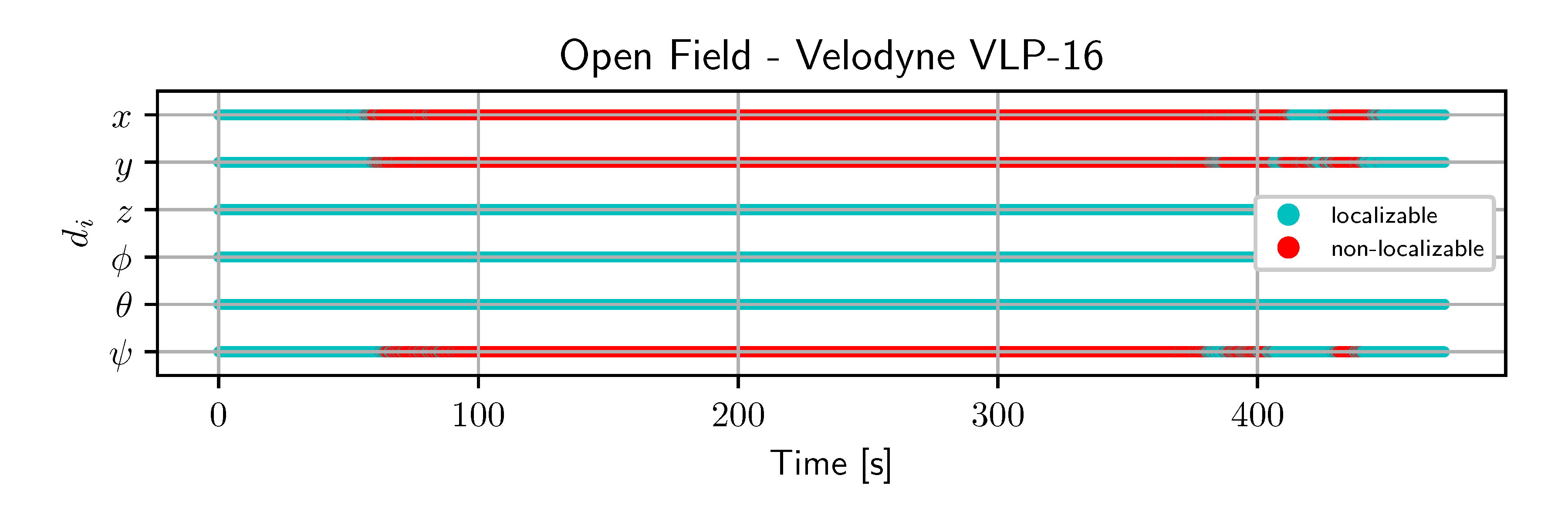}}
\caption{Maps for e-value thresholds of ${20,30}$, and localizability results for the open field experiment. In this scenario, an eigenvalue threshold of 30 is sufficient to detect the degeneracy.}
\label{fig:ruemlang_map}
\vspace{-2ex}
\end{figure}

Utilizing the same experimental setup with the same trained network from the previous section, a field robotic deployment was conducted on an open field near a work site in R\"umlang, Switzerland. During this experiment, the robot starts underneath a canopy, and then traverses over a large concrete field without any geometric features nearby, causing the LiDAR-based scan-to-scan registration to become degenerate along $x$, $y$ and $\psi$ (yaw) directions, as also predicted by the network in lower Figure~\ref{fig:ruemlang_map}. The proposed method can predict the localizability correctly resulting in an accurate map of the environment to be built, as shown in the top of Figure~\ref{fig:ruemlang_map}. Furthermore, it can be noted that a different eigenvalue threshold needs to be set for the degeneracy detection method to work properly as compared to the experiment in the tunnel environment, whereas the same trained network is deployed across both experiments demonstrating the potential of the proposed method to work across different real-world and challenging environments.

\subsubsection{Urban Environment}
\begin{figure}[t!]
\includegraphics[width=1.0\columnwidth]{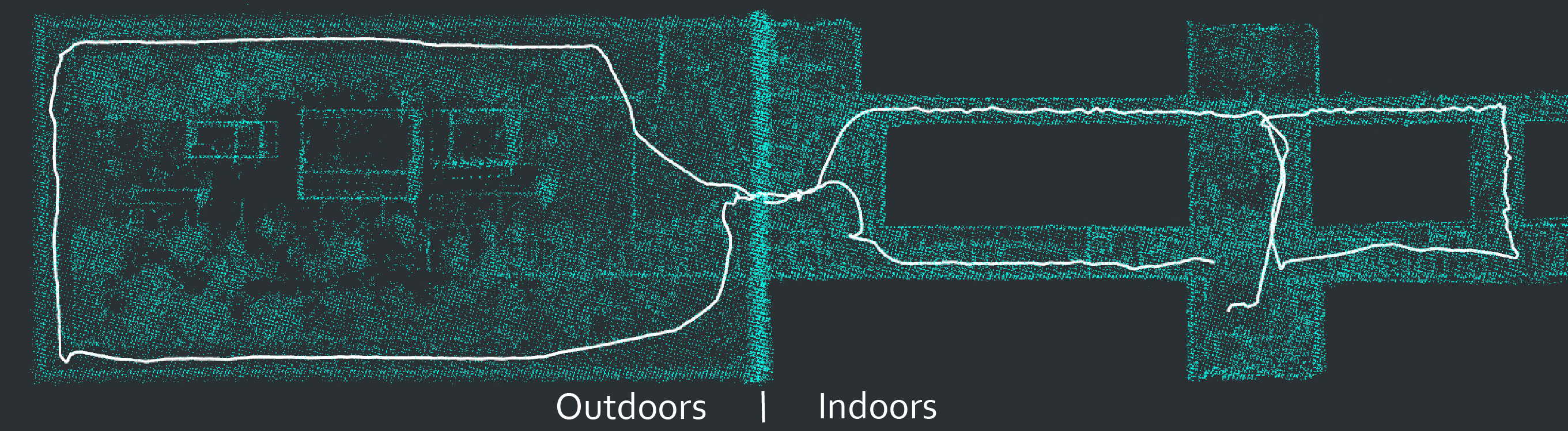}
\adjustbox{trim=0 {0.08\height} 0 0,clip, width=\columnwidth}
{\includegraphics[width=\columnwidth]{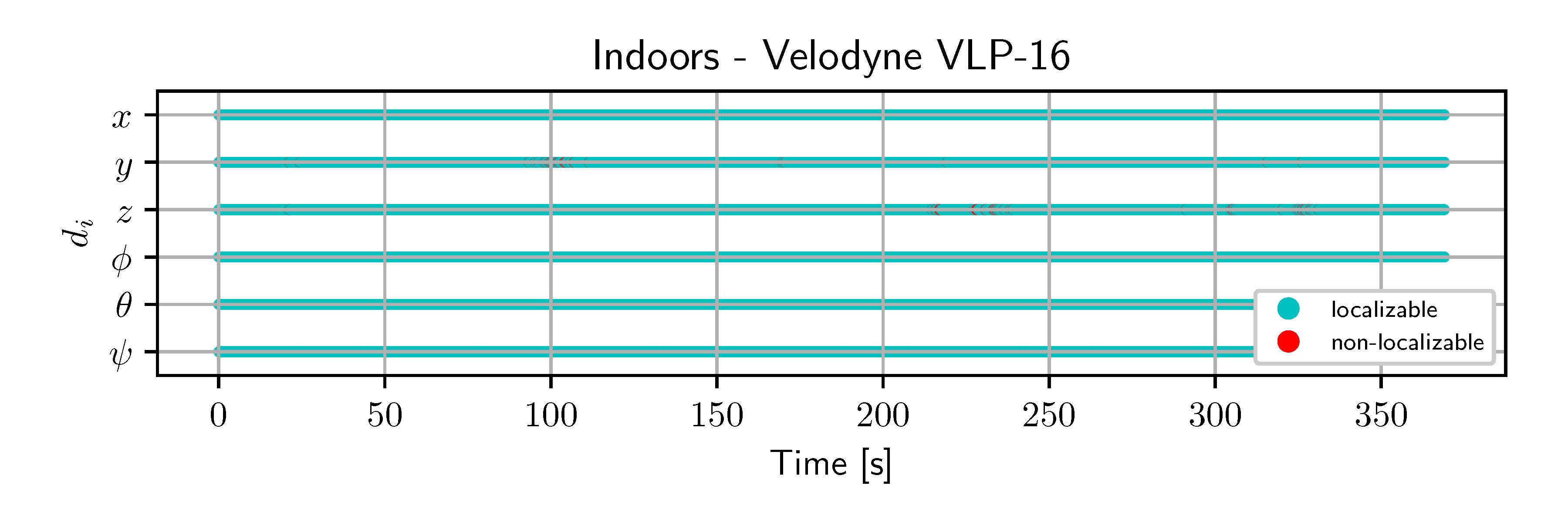}}
\caption{Created map and localizability results in the office environment. Localizability is detected to be present for this scenario, so the full scan-to-scan registration is used as a prior to the mapping.}
\label{fig:eth_map}
\vspace{-2ex}
\end{figure}

Towards evaluating the performance of the proposed method in urban environments, a robotic experiment was conducted in an office-like environment at ETH Z\"urich, Switzerland. This environment contains narrow corridors, glass surfaces and open spaces both in indoor and outdoor settings. During the experiments, the robot starts indoors and traverses outdoors on a roof-top terrace before looping back to the start position. The robot path and the generated environments map can be seen in upper Figure~\ref{fig:eth_map}. In the lower plot it can be noted that the proposed approach provides the correct localizability estimates during this experiment, mostly localizable, validating it as a potential solution for operation across a variety of environments.
\begin{figure}[b!]
\adjustbox{trim=0 {0.08\height} 0 0,clip, width=\columnwidth}
{\includegraphics[width=\columnwidth]{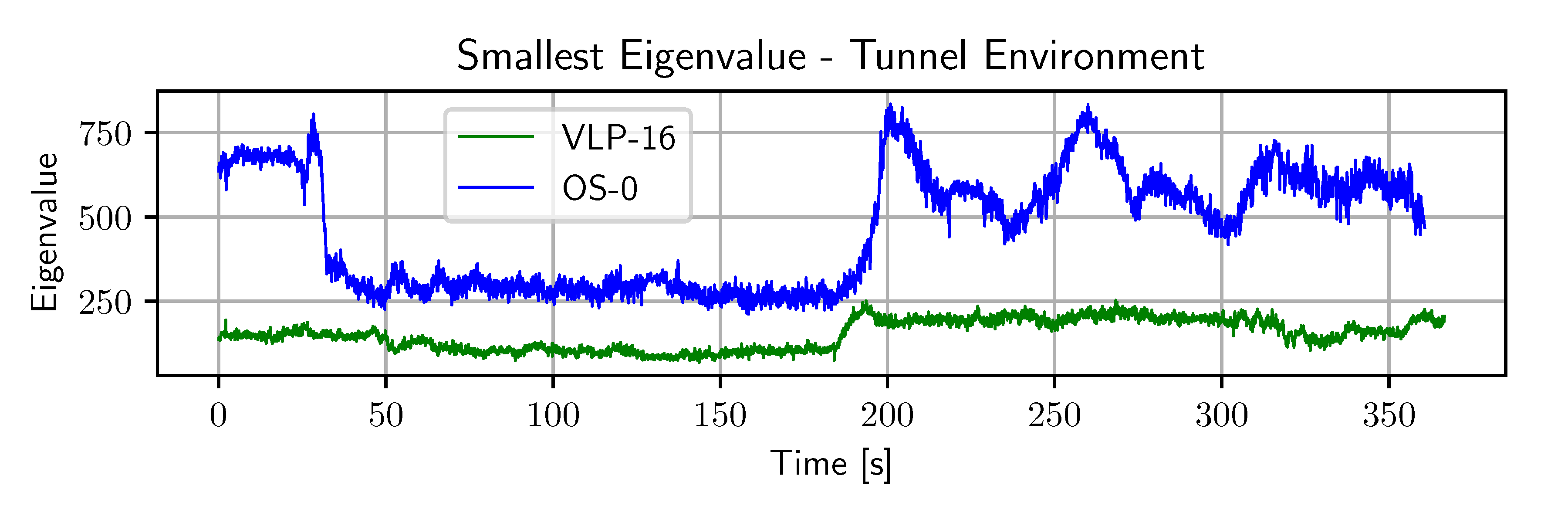}}
\adjustbox{trim=0 {0.08\height} 0 0,clip, width=\columnwidth}
{\includegraphics[width=\columnwidth]{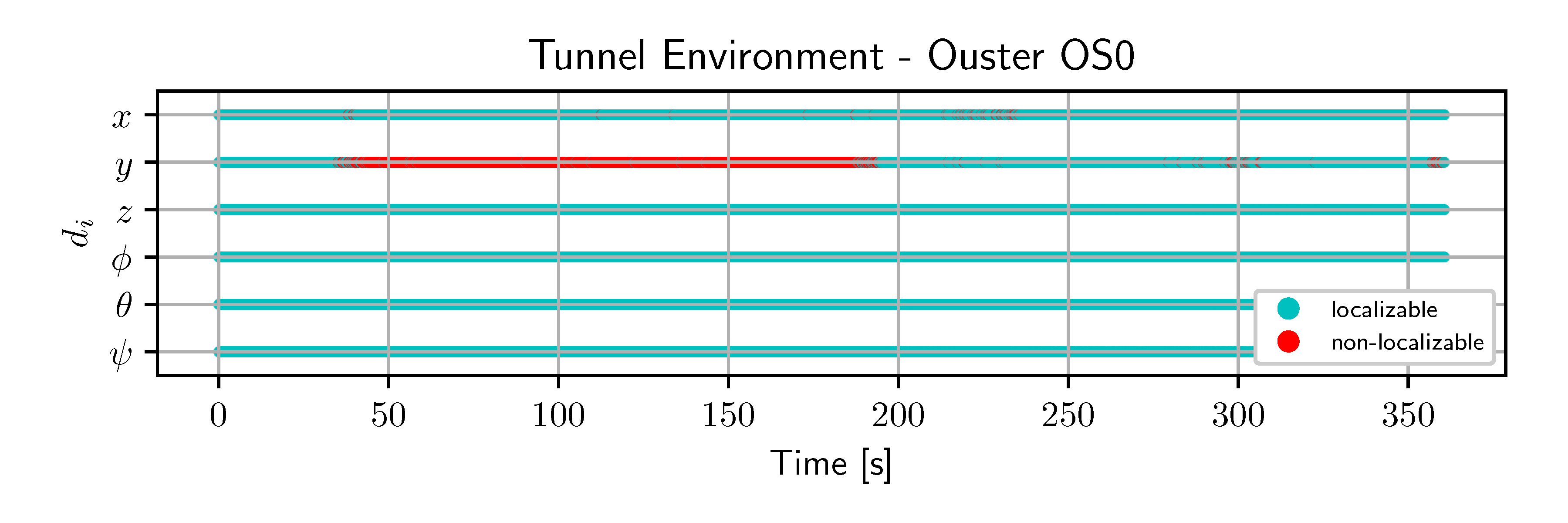}}
\caption{Comparison of the smallest eigenvalues of the detection approach in~\cite{zhang2016degeneracy} for Velodyne VLP-16 and the Ouster OS0-128 LiDAR sensors. Although traversing the same environment, the scale of the eigenvalues differs significantly. Despite this, the proposed network detects non-localizability reliably without any refinement.}
\label{fig:seemuehle-eigenvalues}
\end{figure}

\subsection{Generalization}
To evaluate the generalization capabilities of the proposed approach, experiments are performed with a different LiDAR sensor type, whose measurements have not been observed during training. For this experiment, a similar robot path is traversed through the same underground mine tunnel presented in Section~\ref{sec:tunnel_env}, with the robot being equipped with a different LiDAR sensor; a 128-beam Ouster OS0 LiDAR with a 90 degree vertical field-of-view. This represents a big step from the 16-beam Velodyne VLP-16 sensor with only a 30 degree vertical field-of-view that was used during training and previous experiments.
This sensor variation imposes a large change on the environment observation during a single scan and a significant increase in the input point cloud data. To illustrate the difference quantitatively, the smallest eigenvalues, i.e. the ones traditionally used for degeneracy detection, are shown for both sensor types along the path in upper Figure~\ref{fig:seemuehle-eigenvalues}: The scale of the optimization problem is impacted significantly, necessitating heuristic threshold to be re-tuned for this experiment. However, comparing the bottom plots in Figure~\ref{fig:seemuehle_maps} and Figure~\ref{fig:seemuehle-eigenvalues}, one can conclude that the proposed approach's estimation ability is not significantly impacted by the sensor change, and is still able to correctly predict localizability for both sensor types.

\section{Conclusions \& Future Work}\label{sec:conclusions}
This work presented an end-to-end trained approach for localizability estimation of LiDAR-based point cloud registration. The presented approach is purely trained on simulated data, allowing for an easy scale-up of the generated training data, if needed. Suitability of the selected architecture and training procedure is shown through an ablation study. Practicality and the ability to generalize to real world robotic use-cases is demonstrated through three experiments conducted in i) a self-symmetric tunnel and cave mission, ii) a large scale planar outdoor scenery, and iii) in an indoor office environment.
The same trained network was deployed across all experiments without the requirement of heuristic threshold tuning. 
For future work, one promising direction of research is to extend the presented approach to predict a full 6-DOF covariance matrix $\mathcal{M} \in \mathbb{R}^{6 \times 6}$, which would allow for precise localizability estimation in scenarios where the robot is misaligned with the environment. Further plans include making use of this degeneracy information in the scan-to-map registration process as well. Finally, localizability detection of different sensor modalities is expected to help towards more robust and reliable sensor fusion, e.g. in the form of partial factors in optimization-based approaches~\cite{kaess2019degeneracy,nubert2022graph}.

\addtolength{\textheight}{-12cm}


\bibliographystyle{IEEEtran}
\bibliography{RAL2022}

\end{document}